%% file: iros2021_yoshimura.tex
\documentclass[twocolumn,conference, letterpaper]{ieeeconf}
\usepackage[T1]{fontenc}
\usepackage[latin9]{inputenc}
\usepackage{color}
\usepackage{array}
\usepackage{multirow}
\usepackage{amsmath}
\usepackage{amssymb}
\usepackage{tablefootnote}
\usepackage[unicode=true,
 bookmarks=true,bookmarksnumbered=true,bookmarksopen=true,bookmarksopenlevel=1,
 breaklinks=false,pdfborder={0 0 0},pdfborderstyle={},backref=false,colorlinks=false]
 {hyperref}
\hypersetup{pdftitle={Your Title},
 pdfauthor={Your Name},
 pdfpagelayout=OneColumn, pdfnewwindow=true, pdfstartview=XYZ, plainpages=false}

\makeatletter

\providecommand{\tabularnewline}{\\}
\usepackage{tikz}
\usetikzlibrary{calc}

\usepackage[caption=false,font=footnotesize]{subfig}

\usepackage{graphicx}
\usepackage{import}

\usepackage{cite}

\usepackage{balance}

\usepackage{resizegather}

\IEEEoverridecommandlockouts 

\makeatother

\begin{document}
\title{MBAPose: Mask and Bounding-Box Aware Pose Estimation of Surgical Instruments
with Photorealistic Domain Randomization}
\author{Masakazu~Yoshimura\emph{,} Murilo~M.~Marinho, Kanako~Harada, Mamoru~Mitsuishi\thanks{This
work was supported by JSPS KAKENHI Grant Number JP19K14935. \emph{(Corresponding
author:} Murilo M. Marinho)}\thanks{Masakazu Yoshimura, Murilo M.
Marinho, Kanako Harada, and Mamoru Mitsuishi are with the Department
of Mechanical Engineering, the University of Tokyo, Tokyo, Japan.
\texttt{Emails:\{m.yoshimura, murilo, kanako, mamoru\}@nml.t.u-tokyo.ac.jp}.
}}
\maketitle
\begin{abstract}
Surgical robots are usually controlled using a priori models based
on the robots' geometric parameters, which are calibrated before the
surgical procedure. One of the challenges in using robots in real
surgical settings is that those parameters can change over time, consequently
deteriorating control accuracy. In this context, our group has been
investigating online calibration strategies without added sensors.
In one step toward that goal, we have developed an algorithm to estimate
the pose of the instruments' shafts in endoscopic images. In this
study, we build upon that earlier work and propose a new framework
to more precisely estimate the pose of a rigid surgical instrument.
Our strategy is based on a novel pose estimation model called MBAPose
and the use of synthetic training data. Our experiments demonstrated
an improvement of 21 \% for translation error and 26 \% for orientation
error on synthetic test data with respect to our previous work. Results
with real test data provide a baseline for further research.
\end{abstract}

\section{Introduction}

Robot-assisted surgery is one of the promising contributions of robotics
to the medical field. Robots enable surgeons to perform complex manipulations
with robotic instruments that have multiple degrees of freedom at
the tip. In this context, we have developed and validated a teleoperated
surgical robot, SmartArm \cite{Marinho2020}, with the aim of conducting
procedures in constrained workspaces.

One of the applications of the SmartArm system is endoscopic transsphenoidal
surgery, wherein instruments are inserted through the nostrils to
remove tumors in the pituitary gland or at the base of the skull.
Owing to the narrow endonasal workspace, suturing the dura mater after
tumor removal is especially difficult and this task can highly benefit
from robotic aid. We have developed a control strategy based on the
kinematic model of the robots to make suturing in a narrow workspace
possible by automatically preventing collisions between robots, instruments,
and the patient \cite{Marinho2019,marinho2019dynamic}.

Even with a careful offline calibration of the robots' parameters,
a mismatch of a few millimeters between the kinematically calculated
pose\footnote{Pose here indicates combined position and orientation.}
and the real pose of the instruments. This mismatch tends to increase
during surgery, for instance, owing to the changes in temperature
and interactions among the instruments and tissues. This mismatch
is also reported in related literature \cite{reiter2014appearance,Moccia2020}
with the da Vinci Surgical System (Intuitive Surgical, USA), which
is a commercially used surgical robot.

Aiming for an online calibration strategy, our group is investigating
the use of synthetic images for training deep-learning algorithms
to consistently address the requirements of data-hungry algorithms
that currently dominate state-of-the-art object pose detection. We
previously proposed a pose estimation method for the instruments'
shafts that uses monocular endoscopic images \cite{yoshimura2020single},
in which the model was trained and validated with synthetic images.
In another study, we have investigated physically-based rendering
(PBR) as a means of creating photorealistic images \cite{Perez2020}.
We herein build upon \cite{yoshimura2020single,Perez2020} and propose
a novel framework for the pose estimation of rigid instruments with
PBR-rendered images. Moreover, an improved pose estimation model is
proposed.

\subsection{Related Works}

\begin{figure}[t]
\centering

\def\svgwidth{1.0\columnwidth}

\scriptsize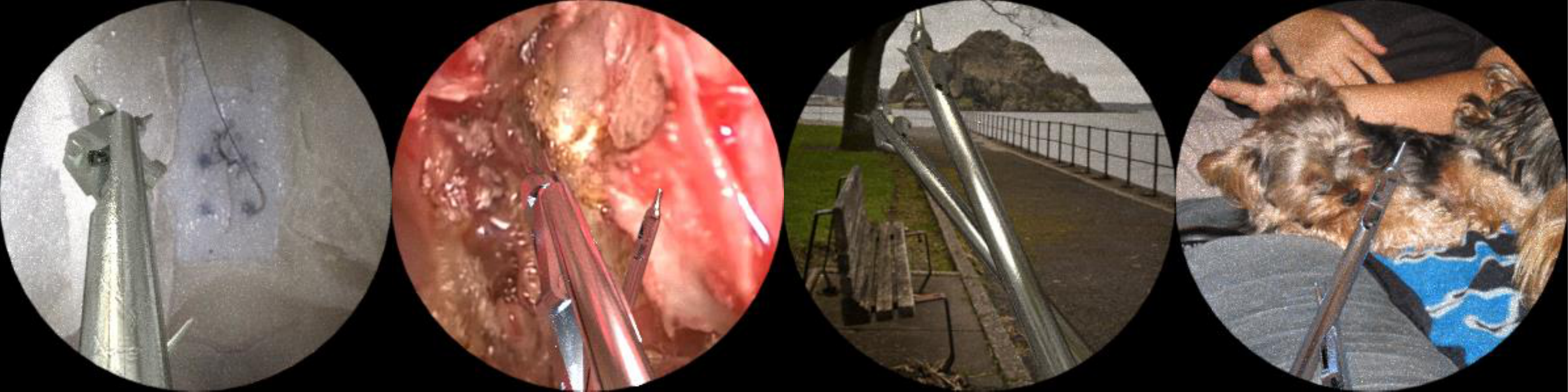

\caption{\label{fig:proposed_dataset} Our proposed dataset that blends photorealistic
rendering and domain randomization optimized for endoscopic views.}
\end{figure}

The biggest challenge in the use of synthetic images for training
deep-learning algorithms is the so-called \emph{domain gap}. In the
context of this work, domain gap is considered as the difference in
the performance of the algorithm trained with only synthetic images
when it is evaluated against real images. Bridging the domain gap
is an open research problem and is being investigated on many fronts,
such as in improved rendering, domain adaptation, and domain randomization
strategies.

\subsubsection{Rendering}

One of the solutions for the domain gap is to render photorealistic
images. To that end, PBR has been used in many contexts \cite{gaidon2016virtual,wrenninge2018synscapes,hodavn2019photorealistic,Perez2020,proencca2020deep}.
PBR yields visually plausible images by solving physics-inspired equations.
Nonetheless, creating 3D models for all elements in the workspace
can be time-consuming, especially when a large variety of images are
required for training.

To partially address this difficulty, the \emph{cut-and-paste }method
\cite{dwibedi2017cut} was proposed. Synthetic image variety increases
by cutting rendered objects and pasting them over real background
images. Many works have used this strategy \cite{kehl2017ssd,tremblay2018training,loquercio2019deep,yoshimura2020single,lee2020camera},
but its adaptivity to real images is relatively low when compared
to PBR. Our prior work \cite{Perez2020} has also provided evidence
of this in a surgical-robot context.

\subsubsection{Domain Adaptation}

Other works \cite{saito2019strong,hsu2020progressive,hoffman2018cycada,su2020adapting}
have proposed domain adaptation methods in which the style of the
real image is transferred to the synthetic image using adversarial
training methods \cite{goodfellow2014generative}. With the state-of-the-art
domain adaptation method for car detection \cite{su2020adapting},
the average precision metric was further improved by 4.6\% when compared
to PBR. However, domain adaptation methods might be unstable and sub-optimal
depending on the dataset \cite{hsu2020progressive}.

\subsubsection{Domain Randomization}

Another method to compensate for the domain gap is a type of data
augmentation strategy called domain randomization (DR) \cite{tobin2017domain}.
The concept behind DR is to widen the synthetic data domain by adding
randomized variations possibly without contextual plausibility or
photorealism. DR has been used, for instance, in robot picking \cite{tobin2017domain},
robot pose estimation \cite{lee2020camera}, and destination detection
for drone racing \cite{loquercio2019deep}. The quality of algorithms
trained with DR is competitive to PBR in some contexts, despite using
cut-and-paste rendering \cite{dwibedi2017cut}. The major interest
in DR is to know what type of randomization is effective in a particular
context. In previous methods, the following characteristics have been
investigated: intensity of lights \cite{tremblay2018training,tobin2017domain,lee2020camera,loquercio2019deep},
position of lights \cite{tremblay2018training,tobin2017domain,lee2020camera},
number of lights \cite{tremblay2018training,tobin2017domain}, type
of lights \cite{tremblay2018training}, object textures \cite{tremblay2018training,tobin2017domain,loquercio2019deep},
shape of the target objects \cite{loquercio2019deep}, addition of
distractors \cite{tobin2017domain,tremblay2018training,loquercio2019deep,lee2020camera},
and randomized background images for cut-and-paste rendering \cite{lee2020camera,tremblay2018training,loquercio2019deep}.

\subsubsection{Pose Estimation}

Several studies have proposed instrument pose estimation using deep
learning. Most existing methods are two-stage. They estimate 2D information
such as key points \cite{peng2019pvnet}, or projected corner points
of 3D bounding boxes \cite{bukschat2020efficientpose}, in the first
stage and subsequently use an optimization algorithm in the second
stage using the information of the first stage to obtain the instrument's
pose. Our strategy is a single-stage and directly predicts the pose
of the surgical instrument.

\subsection{Statement of Contributions}

In this work, we build upon our previous work \cite{yoshimura2020single}
and improve from the pose estimation of the instrument's shaft to
pose estimation of the instruments' tips. Moreover, we (1) improved
the PBR rendering strategy, (2) studied the effects of different DR
strategies and proposed two new randomized components, (3) used an
improved pose estimation model, and (4) discussed how to practically
obtain a real instrument pose dataset and provided results of validation
using real data.

To the best of our knowledge, this is the first work to combine those
strategies and validate them in the pose estimation of rigid surgical
instruments.

\section{Problem Statement}

\begin{figure}[h]
\centering

\def\svgwidth{1.0\columnwidth}

\scriptsize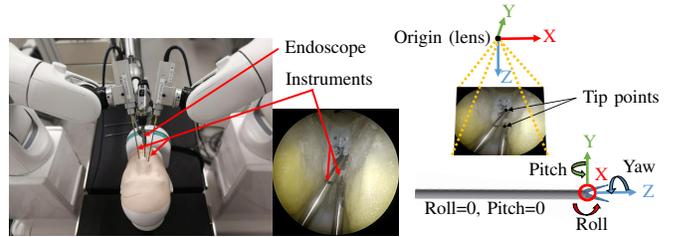

\caption{\label{fig:statement} Robot setup (left), its inner view from a 70$^{\circ}$
endoscope (middle), and the definition of the pose (right).}
\end{figure}

The target application for our method is robotic-assisted endoscopic
transsphenoidal surgery, as shown in Fig.~\ref{fig:statement}. Due
for practical reasons, we attached one 3~mm 2-degrees-of-freedom
(2-DoF) forceps to each robotic arm (in contrast with the hollow shafts
used in \cite{yoshimura2020single}). Those forceps are inserted through
the nostrils of an anatomically realistic head model (BionicBrain
\cite{Masuda2019}). The workspace is viewed through a 70$^{\circ}$
endoscope (Endoarm, Olympus, Japan) whose perspective angle is 95$^{\circ}$.

In this study, our target is to estimate the pose of the instruments
relative to the endoscope. The reference frames of the instruments
are defined at the distal end of the instrument's shaft and the $z-$axis
is along the instruments' shaft.

The approximate range-of-motion of the instruments in the endonasal
workspace is listed in Table~\ref{tab:dataset_range}.

\begin{table}[tbh]
\centering

\caption{\label{tab:dataset_range} Possible range of the instrument pose relative
to the endoscope.}

\noindent\resizebox{\columnwidth}{!}{%

\textcolor{black}{}%
\begin{tabular}{cccccccc}
\noalign{\vskip\doublerulesep}
Dimension & x {[}mm{]} & y {[}mm{]} & z {[}mm{]} & roll {[}degree{]} & pitch {[}degree{]} & yaw {[}degree{]} & gripper {[}degree{]}\tabularnewline[\doublerulesep]
\hline 
\noalign{\vskip\doublerulesep}
\noalign{\vskip\doublerulesep}
Range & -20\textasciitilde 20 & -20\textasciitilde 20 & 10\textasciitilde 35 & 50\textasciitilde 90 & -40\textasciitilde 40 & 0\textasciitilde 360 & 0\textasciitilde 60\tabularnewline[\doublerulesep]
\noalign{\vskip\doublerulesep}
\end{tabular}

}
\end{table}

\section{Methodology}

Our pose estimation strategy is based on three main points-{}-{}-a
realistic rendering strategy, a DR strategy, and a single-stage pose
estimation model.

\subsection{Proposed realistic rendering strategy}

\begin{figure}[h]
\centering

\def\svgwidth{1.0\columnwidth}

\scriptsize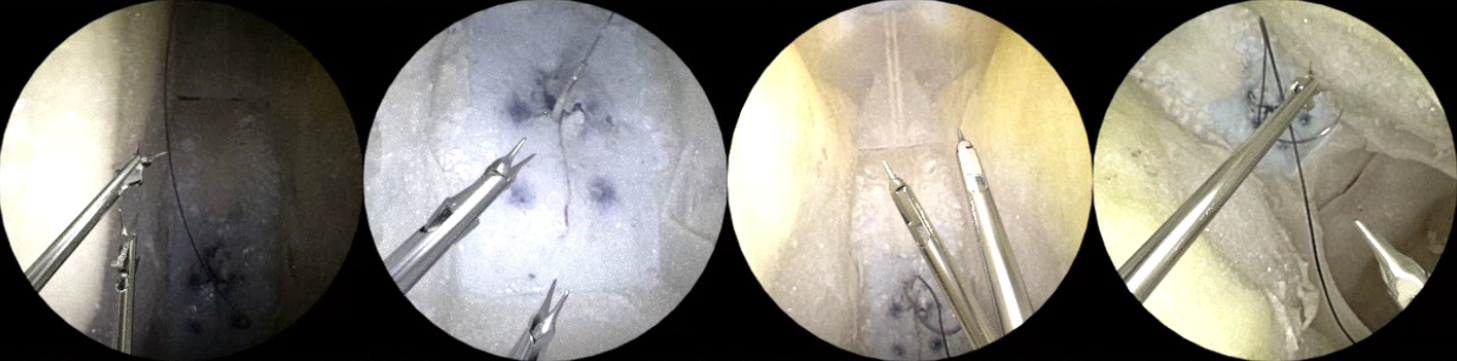

\caption{\label{fig:sample_artificial_images} Samples rendered with the proposed
realistic rendering strategy.}
\end{figure}

The proposed realistic rendering strategy uses elements of PBR and
the cut-and-paste method. It calculates light reflection, diffusion,
and absorption faithfully with PBR using ray tracing \cite{kajiya1986rendering}
and the concept of microfacet \cite{cook1982reflectance} n the same
manner as in previous PBR methods \cite{wrenninge2018synscapes,gaidon2016virtual,hodavn2019photorealistic,proencca2020deep}.
The background is not rendered using PRB; instead, as described in
our previous works \cite{yoshimura2020single}, we render a cube whose
faces are real background images. The reflection of the surroundings
in the instruments are rendered by tracing photons between the cubic
faces and the instruments' surface.

In a direct comparison with \cite{yoshimura2020single}, we improved
several rendering methods. First, we applied a bidirectional scattering
diffusion function (BSDF) with Oren\textendash Nayar reflectance model
\cite{oren1994generalization} on the cubic surface to reflect it
in various directions. This was necessary because we simplified the
surroundings as a cube. A BSDF with GGX\footnote{GGX itself the name of the distribution and it does not seem to be
any particular acronym \cite{walter2007microfacet}.} microfacet distribution \cite{walter2007microfacet} was used on
the instruments' surface to obtain realistic reflection. We positioned
a black plane with a hole in front of the camera to reproduce the
circular endoscopic view. The sheet was small and placed close to
the camera so as to not interfere with the ray tracing.

The examples of synthetic images created with the proposed rendering
method are shown is Fig.~\ref{fig:sample_artificial_images}.

\subsection{Domain Randomization}

\begin{figure*}[tbh]
\centering

\def\svgwidth{2.0\columnwidth}

\scriptsize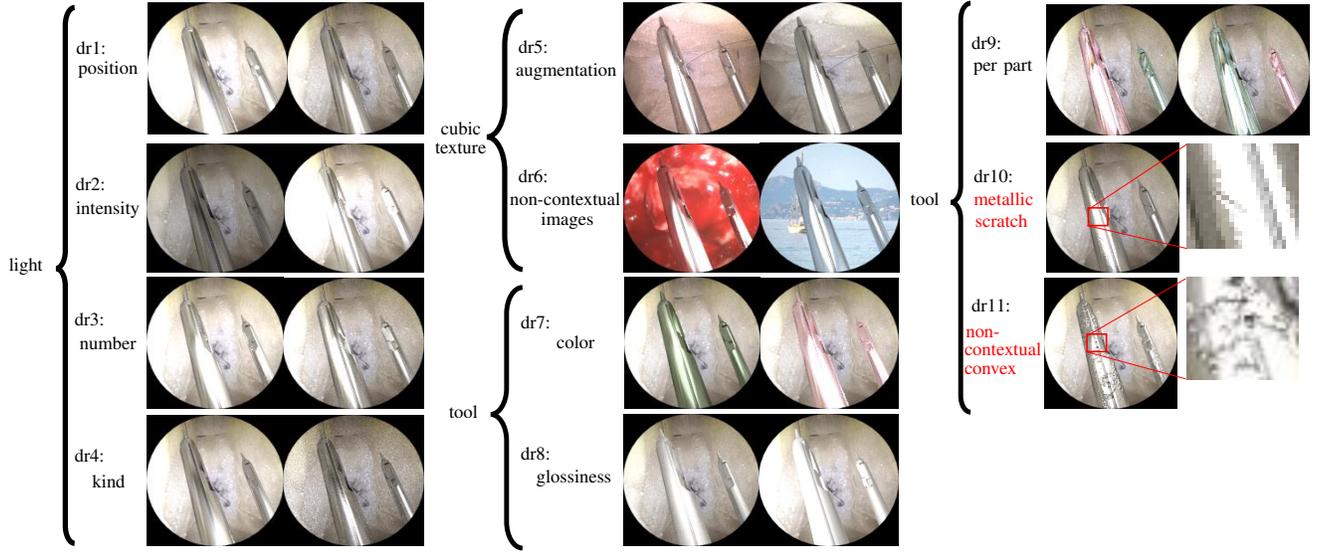

\caption{\label{fig:dr_compnents} Effect of DR components. The components
marked red are new methods.}
\end{figure*}

DR methods have usually been applied together with classic or simple
rendering. We herein combine PBR with DR and propose two further DR
strategies.

We use the following previously propose DR strategies: we randomize
the position, intensity, number, and type of lights. In addition,
the real background images used for the cubic texture are randomly
selected. For the background images, we separately evaluated only
the real images of the experimental environment, images of the experimental
environment and their augmented images, or these plus non-contextual
images. As for the surgical instruments, we subdivided the previously
proposed texture randomization \cite{tobin2017domain} into color
randomization and glossiness randomization for the metallic surface.
We were also interested in analyzing the effect of randomizing the
textures for each individual instrument separately.

In addition to those well-known DR strategies, we propose a further
surface randomization method. It randomizes the appearance instrument's
surface using normal mapping \cite{cignoni1998general}. Normal mapping
is a strategy used to change surface appearance of objects by changing
the normal vectors of the surface at the rendering time. We randomly
created concave and convex surface imperfections by changing the normal
maps.

The performance of each strategy in the context of this work is evaluated
in the experiments in comprehensive ablation studies.

\subsection{Pose Estimation Model}

\begin{figure}[h]
\centering

\def\svgwidth{1.0\columnwidth}

\scriptsize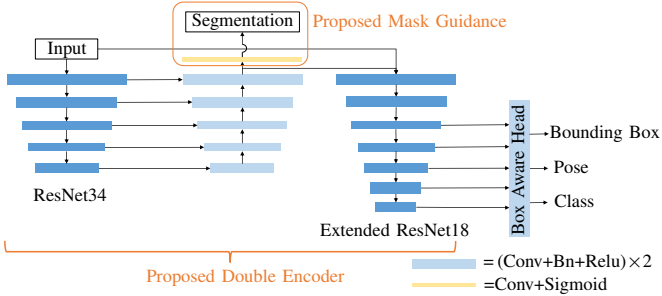

\caption{\label{fig:model} Proposed Model Architecture. The Box-Aware Head
is the Architecture-D model defined in \cite{yoshimura2020single}.}
\end{figure}

We propose a new pose estimation model named MBAPose to predict the
pose of instruments is aware of their mask and bounding box shape.
This model is a combination of two parts, a prediction head and a
feature extraction model.

The prediction head, which we call the Box-Aware Head, is one we proposed
in \cite{yoshimura2020single}. It uses the features obtained from
the feature extraction model to simultaneously predict the bounding
box, class, and pose. The architecture that internally used self-bounding
box prediction and self-class prediction to predict their pose enhanced
pose estimation. .

This feature extraction is novel and composed of a double encoder
model with mask guidance. The double encoder model itself comprises
two smaller encoder models and one decoder model. We propose this
feature extractor because our previous model \cite{yoshimura2020single}
had difficulties in detecting small objects, as shown in \cite{yoshimura2020on}.
This change was inspired by DSSD \cite{fu2017dssd}, that improved
the detection of small objects by adding a decoder network. Different
from DSSD, we further add a second encoder. The aim is to create a
mask-aware feature by adding a loss function with an additional convolution
at the head of the decoder model in the training stage. By adding
segmentation loss, the model estimates the pose of the objects aware
of their contour. With the proposed double encoder, the contour information
can be more explicitly used compared with others \cite{hu2019segmentation}.

\subsection{Test dataset with real instrument's image and pose information\label{subsec:test_dataset}}

\begin{figure}[h]
\centering

\def\svgwidth{0.9\columnwidth}

\scriptsize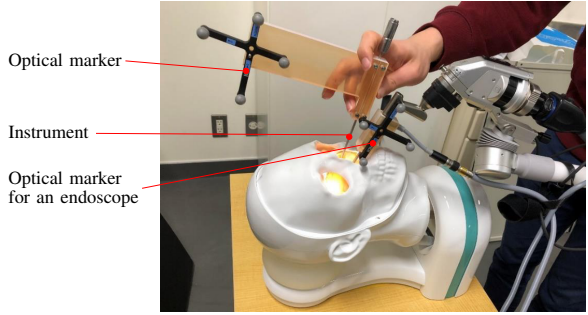

\caption{\label{fig:polaris} Setup for test dataset acquisition using an optical
tracker.}
\end{figure}

One challenge we faced in our prior work \cite{yoshimura2020single}
was the difficulty in creating a reliable dataset for testing the
model with the image and pose information of the real instruments.
In this work, we overcame those difficulties and created a test dataset
as follows.

The pose of the real instruments was acquired using an optical tracker
(Polaris Vega, NDI, Canada) attached to its optical marker at the
base of the instruments and endoscope, as shown in Fig.~\ref{fig:polaris}
using 3D-printed attachments. However, even small mounting and fabrication
errors induce a considerably large error at the tip of the instrument.
To calibrate for the mounting error, we propose a practical markerless
hand-eye calibration method using the contour of instruments in the
view. The method is as follows.

The present hand-eye calibration problem can be represented using
homogenous-transformation matrices as $\boldsymbol{AX}=\boldsymbol{ZCB}$,
where $\boldsymbol{A}$, $\boldsymbol{B}$, $\boldsymbol{C}$, $\boldsymbol{X}$,
and $\boldsymbol{Z}$ denote the transformation from the instrument's
tip pose to the endoscopic lens pose, the instrument's maker pose
with respect to the world, the endoscope's marker pose with respect
to the world, the transformation from the instruments' maker pose
to the instrument's tip pose, and the transformation from the endoscope's
maker pose to the endoscopic lens pose. $\boldsymbol{B}$ and $\boldsymbol{C}$
are acquired from the optical tracker. We have the ideal values for
$\boldsymbol{X}$ and $\boldsymbol{Z}$ using our CAD models; however,
they are corrupted by the mounting errors and consequently our calibration
target. $\boldsymbol{A}$ is usually estimated with a checkerboard;
however, in our case, we do not obtain $\boldsymbol{A}$ explicitly
because installing a checkerboard pattern on the tip of the instrument
is not feasible. Instead, we indirectly obtain $\boldsymbol{A}$ using
the contour of instruments on the images.

\begin{figure}[h]
\centering

\def\svgwidth{1.0\columnwidth}

\scriptsize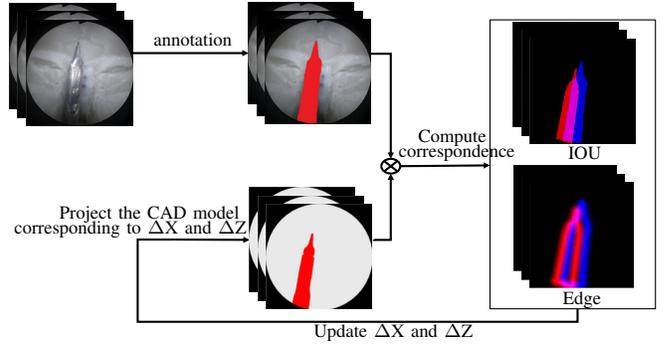

\caption{\label{fig:hand_eye} The proposed hand-eye calibration method.}
\end{figure}

We search $\boldsymbol{X}$ and $\boldsymbol{Z}$ by iteratively comparing
manually annotated contours on the real images and contours of projected
3D CAD models based on our current estimated of $\boldsymbol{X}$
and $\boldsymbol{Z}$. To do so, we use the following objective function
\begin{equation}
O=\sum_{n}^{N}\left\{ \alpha O_{iou}\left(x_{n}^{gt},x_{n}^{proj}\right)+\left(1-\alpha\right)O_{edge}\left(x_{n}^{gt},x_{n}^{proj}\right)\right\} ,\label{eq:overlap}
\end{equation}
where $x_{n}^{gt}$ and $x_{n}^{proj}$ are the $n-$th contours on
the ground truth and projected CAD model. In detail, $O_{iou}$ is
intersection over union (IoU). In addition, $O_{edge}$ is a metric
that quantifies edge overlap as
\[
O_{edge}\left(x_{gt},x_{proj}\right)=\frac{\sum_{u}\sum_{v}p_{gt}\left(u,v\right)p_{proj}\left(u,v\right)}{\sum_{u}\sum_{v}p_{gt}\left(u,v\right)^{2}},
\]
where $p(u,v)$ are the distance transformed pixel values defined
\[
p\left(u,v\right)=\left(d_{max}-max\left(d_{nearest}\left(u,v\right),d_{max}\right)\right)^{2},
\]
where $d_{nearest}\left(u,v\right)$ is the distance from the nearest
edge, and $d_{max}$ is the maximum distance. With this heuristic,
we search for, $\boldsymbol{X}$ and $\boldsymbol{Z}$, to maximize
$O$ using a ternary search, as shown in Fig.~\ref{fig:hand_eye}.

\section{Evaluation}

First, we evaluate the effect of the proposed hand-{}-eye calibration
to create test data with the improved ground truth of the real pose.
Then, we investigate each component of the DR method and the proposed
rendering method. Third, we evaluate the proposed MBANet. Finally,
the pose estimation performance using all proposed methods is reported.

\subsection{Evaluation of the Proposed Hand-Eye Calibration}

In this section, we briefly test the performance of our practical
hand-eye calibration.

First, we prepared three sequences of endoscopic images paired with
the optical tracker's pose estimation synchronized at a rate of 20~Hz.
In each scene, the instrument was manually moved in the head model.
Only one instrument was used at this stage because the attachments
and markers make it impossible to simultaneously insert a second instrument.

We calibrated the endoscopic camera with the method proposed by \textsl{Zhang
et al}. \cite{zhang2000flexible}. A total of 5301 data pairs were
prepared. Given that the contours have to be manually annotated, we
used 20 images from two sequences to estimate the mounting error.
The hyperparameters $\alpha$ and $d_{max}$ were set as 0.8 and 10,
respectively. The mounting errors were searched in the interval $\left[-1,1\right]$~mm
for each translation dimension and $\left[-1,1\right]$~degree for
each orientation dimension. These are reasonable boundaries for our
fabrication and mounting accuracy.

We tested with another 20 manually annotated images from the unused
sequence using the 2D projection overlap metric \eqref{eq:overlap}.
The results are summarized in Table~\ref{tab:calib}. The results
on the test images suggest that the mounting errors decreased owing
to our method. The maximum estimated mounting error was 0.6~mm and
0.5~degree. Nonetheless, there was still around 2-3~mm of error
between the projections. We believe the remaining error has two sources.
First, ternary search cannot always obtain the optimal value and in
our case depends on hyperparameters. Second, the visual tracker itself
has an associated accuracy and our best estimates can only be as good
as the tracker's. The tracker can estimate the position of each passive
marker with a 0.12~mm RMS error, that error is amplified by assembly
and fabrication errors in the adapter between the marker and the tip
of the instrument or endoscope.

\begin{table}[tbh]
\centering

\caption{\label{tab:calib} 2D projection overlapping metric \eqref{eq:overlap}
with and without the proposed hand-eye calibration.}

\textcolor{black}{}%
\begin{tabular}{ccc}
Used for optimization & Yes & No\tabularnewline
\hline 
\hline 
Without calibration & 0.33 & 0.47\tabularnewline
With calibration & \textbf{0.71} & \textbf{0.59}\tabularnewline
\end{tabular}

\medskip{}

{\small{}{*}Larger values mean better accuracy.}{\small\par}
\end{table}

\subsection{Dataset Creation}

\subsubsection{Training Dataset\label{subsec:Training-Dataset}}

We rendered 10,000 images for each DR component using an open-source
rendering software Blender (Blender Foundation, Netherlands). The
datasets were rendered not as sequences but as independent scenes
with totally randomized poses. As we fixed the random seeds, all rendered
instruments had the same poses over each specific dataset. Therefore,
any difference in performance between DR cases only stemmed from the
DR component itself. Even if no DR components were used, and 754 contextual
images of the head model (Fig.~\ref{fig:statement}) were used as
textures for the cube. When non-contextual images were used as texture,
the 754 contextual images were used. Along with it, a total of 2840
real endoscopic transsphenoidal surgery scenes, images from the COCO
dataset \cite{lin2014microsoft}, and 754 contextual images images
were used with a ratio of 0.25, 0.5, and 0.25, respectively. Other
randomized parameters are listed in Table~\ref{tab:dr_param}. The
image size was $299\times299$.

\subsubsection{Test Dataset}

The 5301 images in our calibrated real dataset (described in Section~\eqref{subsec:test_dataset})
were used for testing.

\subsection{Evaluation Metrics}

We used translation error, centerline angle error, mean average precision
(mAP) and the detected rate as the performance metrics. The centerline
angle error is obtained from the angle between the estimate instrument
shaft centerline and the real one.

\subsection{Ablation study for the DR components\label{subsec:Evaluation-of-Each}}

In this section, the effect of each DR component is evaluated in our
target application. We trained the model proposed in \cite{yoshimura2020single}
for 120,000 iterations with a batch size of 20 and the one-cycle policy
learning rate \cite{smith2018disciplined} wherein the learning rate
begun at 10$^{-5}$, increased to 10$^{-4}$, and subsequently decreased
to 5$\times$10$^{-7}$.

The pose map has seven channels corresponding to $x$, $y$, and $z$
translations; roll, pitch, and yaw rotations, respectively, and gripper
angle. The pose loss was weighted with empirical factors of 1, 1,
1.5, 2, 2, 0.1, and 0.1.

We used the following data augmentation strategies in all cases: random
contrast, random hue, random saturation, and additive noise augmentation
\cite{yoshimura2020single}. We did not use real images for training
and instead concentrate on the effects of each synthetic dataset.

\begin{figure}[h]
\centering

\def\svgwidth{1.0\columnwidth}

\scriptsize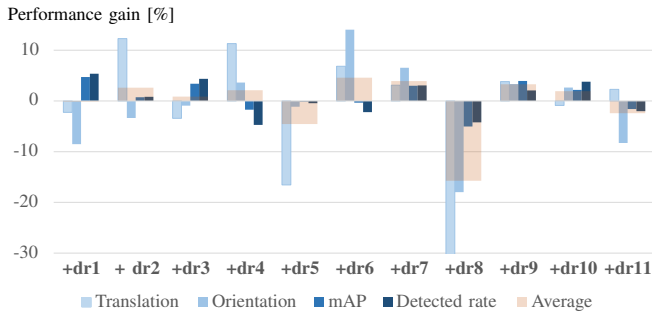

\caption{\label{fig:dr_eval} Ablation study for the DR components. The performance
gain per each metric is shown.}
\end{figure}

The results of this ablation study are summarized in Fig.~\ref{fig:dr_eval}.
Interestingly, some DR strategies deteriorated the generalizability
of the model to real images. For instance, the randomization of the
instrument's glossiness strongly deteriorated the generalizability.
We infer that this means that metallic reflection is important for
detecting and estimating the pose. Non-contextual images used as background
textures improved all metrics. In addition, the proposed normal-map-based
random scratch generation improved all metrics, although further randomization
with non-contextual convex and concave deteriorated the performance.

\begin{table}[tbh]
\centering

\caption{\label{tab:dr_param} Randomized detail of each DR component.}

\noindent\resizebox{\columnwidth}{!}{%

\textcolor{black}{}%
\begin{tabular}{ccc}
 & method & randomized detail\tabularnewline
\hline 
\hline 
dr1 & light position & behind the camera\tabularnewline
dr2 & light intensity\tablefootnote{The value of intensity parameter in Blender.} & 5\textasciitilde 1$\times$10$^{6}$\tabularnewline
dr3 & light number & 1\textasciitilde 2\tabularnewline
dr4 & light kind & point, sun, area, hemi\tabularnewline
dr5 & texture aug. & crop, rotate, color jitter\tabularnewline
dr6 & texture non-contextual & ref. \ref{subsec:Training-Dataset}\tabularnewline
dr7 & instrument color & hue, brightness\tabularnewline
dr8 & instrument glossiness & full range\tabularnewline
dr9 & instrument per part & 50\textbackslash\%\tabularnewline
dr10 & instrument scratch & a normal map of scratch and augment\tabularnewline
dr11 & instrument non-contextual convex & add 19 non-contextual normal maps\tabularnewline
\end{tabular}

}
\end{table}

\subsection{Evaluation of the Overall Synthetic Data Generation}

For this evaluation, we selected the DR components that, on average,
improved the metrics in the ablation study. We compared our strategy
with the cut-and-paste method \cite{dwibedi2017cut}, without Poisson
Blending.

The evaluation results were Table~\ref{tab:overall}. The proposed
renderer had better performance compared to the cut-and-paste method.
Ray-traced reflections are important for pose estimation in our context.

Furthermore, by adding the optimized DR components, the detected rate
and translation were improved although rotation error increased. DR
improves the object detection rate. In terms of pose estimation, DR
was only effective if the correct DR components were used.

\begin{table}[tbh]
\centering

\caption{\label{tab:overall} Performance comparison over the real endoscopic
images.}

\noindent\resizebox{\columnwidth}{!}{%

\begin{tabular}{cc|ccc}
 &  & Tran. {[}mm{]} & Orien. {[}degree{]} & Detected rate\tabularnewline
\hline 
\multirow{4}{*}{Real} & Cut and paste \cite{dwibedi2017cut} & 7.19 & 9.35 & 47.95\tabularnewline
 & Our renderer & 5.29 & \textbf{6.23} & 69.67\tabularnewline
 & Our renderer + all DR & 5.88 & 6.64 & \textbf{73.44}\tabularnewline
 & Our renderer + optimal. DR & \textbf{2.80} & 8.04 & \textbf{73.44}\tabularnewline
\end{tabular}

}

Note that the ground-truth of the real data had around 2-3 mm error
even after the hand-eye calibration.
\end{table}

\subsection{Evaluation of The Pose Estimation Model}

The performance of the proposed model was also evaluated. For this
purpose, 10,000 photorealistic synthetic images without DR were used
as test data. In addition, 90,000 synthetic images with all DR components
are used as the training data. Other training settings were the same
as that mentioned in Section~\ref{subsec:Evaluation-of-Each}.

The implementation details of the pose estimation model were as follows.
ResNet34 \cite{he2016resnet} was used as the first encoder. The output
of the decoder had 16 channels, and it was concatenated with the original
input image to input the second encoder. The second encoder was based
on ResNet18 but extended with two additional layers to detect instruments
when they make most of the image, for instance, when they are close
to the endoscope. Each layer comprised two residual blocks with a
channel size of 512. The loss function is based on multi-task loss
defined in \cite{yoshimura2020single} and binary cross-entropy loss
for mask guidance was added, with a factor of 4.0. To compare with
\cite{yoshimura2020single} in which only the shaft was estimated,
the orientation angle error calculation does not consider the $z-$axis
rotation.In this experiment, we also add a rigid transformation augmentation
\cite{yoshimura2020on}, to allow rotation augmentation of the images
paired with a suitable rotation of the paired instrument pose. We
rotated in the range of {[}-15$^{\circ}$, 15$^{\circ}${]} with 80\%
probability. The inference speed was measured with Python 3.6 using
Pytorch 1.4 and CUDA 10.0 on Ubuntu 18.04 with a Titan RTX graphics
card.

The results are summarized in Table~\ref{tab:model}. The proposed
Double Encoder model is fast and accurate. By including mask guidance,
pose estimation was further improved. Moreover, the rigid geometric
augmentation improved results. The 90,000 synthetic images used can
be considered a small number given that  8.6$\times10^{9}$ images
are needed if a training dataset is created at intervals of 1 mm and
2$^{\circ}$. Therefore, the proposed method could estimate the pose
with a relatively coarse interpolation of the training data.

\begin{table}[tbh]
\caption{\label{tab:model}Evaluation of the proposed model architecture against
synthetic test data.}

\noindent\resizebox{\columnwidth}{!}{%

\begin{tabular}{ccccc}
 & Tran. {[}mm{]} & Orien. {[}degree{]} & Detected rate & fps\tabularnewline
\hline 
\hline 
\cite{yoshimura2020single} & 1.11 & 3.58 & 99.66 & 41.0\tabularnewline
+ double encoder & 0.99 & 3.11 & 99.80 & \textbf{79.4}\tabularnewline
+ mask guidance & 0.97 & 3.03 & 99.81 & \textbf{79.4}\tabularnewline
+ rigid geo. aug. \cite{yoshimura2020on} & \textbf{0.88} & \textbf{2.66} & \textbf{99.84} & \textbf{79.4}\tabularnewline
\end{tabular}

}
\end{table}

\subsection{Overall Performance Evaluation}

Finally, we report the best performance with the proposed PBR, optimized
DR, and proposed pose estimation model in the pose estimation of real
images. In total, 90,000 synthetic images with optimized DR were used
as the training data. Moreover, an additional training method using
a small number of pose-free real images \cite{yoshimura2020single}
was added. The model trained with synthetic data was additionally
trained with the same synthetic data and pose-free 50 real images.
The model was additionally trained for 10,000 iterations with each
batch comprising 2 real images and 18 synthetic images. The model
was trained from real images except with the pose loss. To compare
with \cite{yoshimura2020single} in which only the shaft was estimated,
the orientation angle error calculation does not consider the $z-$axis
rotation.

\begin{figure}[tbh]
\centering

\def\svgwidth{1.0\columnwidth}

\scriptsize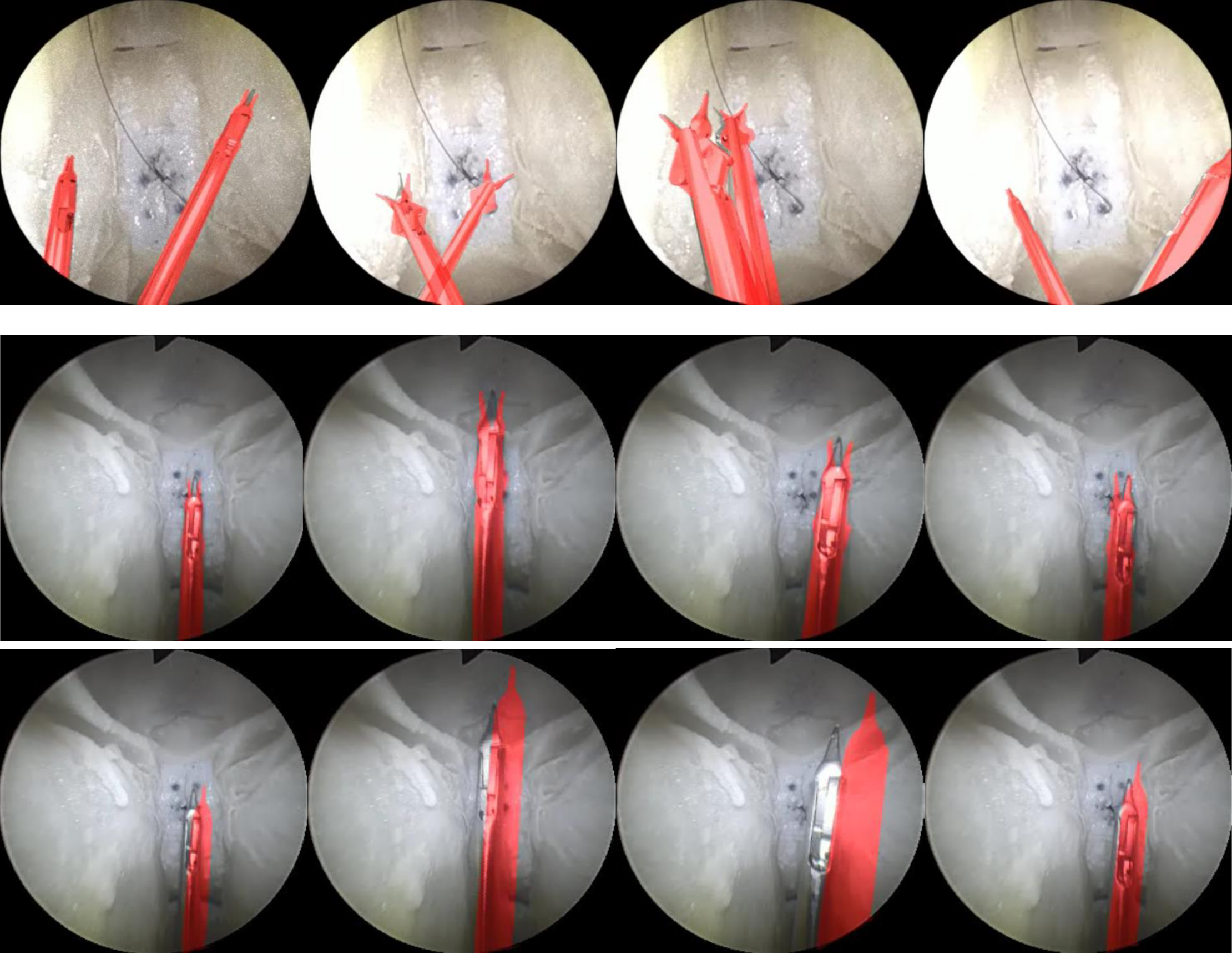

\caption{\label{fig:bounding_box_result} Examples of 2D projections using
the pose estimated by our model. The upper images are synthetic images,
the middle ones are real endoscopic images, and the lower ones are
the corresponding ground truth.}
\end{figure}

The results are summarized in Table~\ref{tab:model}. The pose estimation
improved on synthetic images and real images. In qualitative terms,
the 2D projections of the poses estimated using the proposed method
generally seemed to be more accurate than those of the ground truth
pose as shown in Fig.~\ref{fig:bounding_box_result}. The actual
errors on real images might be smaller than Table~\ref{tab:model},
but we currently do not have a better sensor to test that hypothesis.
The creation of a higher-accuracy real dataset is left for future
work.

\begin{table}[tbh]
\centering

\caption{\label{tab:overall-1} Pose estimation performance by mixing all proposed
methods.}

\noindent\resizebox{\columnwidth}{!}{%

\begin{tabular}{ccccc}
 & Additional training\cite{yoshimura2020single} & Tran. {[}mm{]} & Orien. {[}degree{]} & Detected rate\tabularnewline
\hline 
\hline 
Synthetic &  & 0.84 & 2.50 & 99.84\tabularnewline
\multirow{2}{*}{Real} &  & 2.61 & 7.43 & 74.36\tabularnewline
 & $\checkmark$ & 2.18 & 5.11 & 82.57\tabularnewline
\end{tabular}

}

Note that the ground-truth of the real data had around 2-3 mm error
even after the hand-eye calibration.
\end{table}

\section{Conclusions}

In this paper, we proposed a pose estimation method for surgical instruments.
We achieved a more precise pose estimation than previous works by
combining three methods-{}-a PBR strategy with simplified surroundings,
DR strategies relevant for our particular application, and a pose
estimation network.

In future works, we plan to use sequential video information to increase
the robustness of our predictions. Moreover, we will use the proposed
method in the online calibration of surgical instruments.

\bibliographystyle{IEEEtran}
\bibliography{bib/ral}

\end{document}

%% file: figs/best_cg.pdf_tex
%% Creator: Inkscape inkscape 0.92.1, www.inkscape.org
%% Accompanies image file 'best_cg.pdf' (pdf, eps, ps)
\begingroup%
  \makeatletter%
  \providecommand\color[2][]{%
    \errmessage{(Inkscape) Color is used for the text in Inkscape, but the package 'color.sty' is not loaded}%
    \renewcommand\color[2][]{}%
  }%
  \providecommand\transparent[1]{%
    \errmessage{(Inkscape) Transparency is used (non-zero) for the text in Inkscape, but the package 'transparent.sty' is not loaded}%
    \renewcommand\transparent[1]{}%
  }%
  \providecommand\rotatebox[2]{#2}%
  \ifx\svgwidth\undefined%
    \setlength{\unitlength}{884.64001465bp}%
    \ifx\svgscale\undefined%
      \relax%
    \else%
      \setlength{\unitlength}{\unitlength * \real{\svgscale}}%
    \fi%
  \else%
    \setlength{\unitlength}{\svgwidth}%
  \fi%
  \global\let\svgwidth\undefined%
  \global\let\svgscale\undefined%
  \makeatother%
  \begin{picture}(1,0.25)%
    \put(0,0){\includegraphics[width=\unitlength,page=1]{best_cg.pdf}}%
  \end{picture}%
\endgroup%

%% file: figs/robot2.pdf_tex
%% Creator: Inkscape inkscape 0.92.1, www.inkscape.org
%% Accompanies image file 'robot2.pdf' (pdf, eps, ps)
\begingroup%
  \makeatletter%
  \providecommand\color[2][]{%
    \errmessage{(Inkscape) Color is used for the text in Inkscape, but the package 'color.sty' is not loaded}%
    \renewcommand\color[2][]{}%
  }%
  \providecommand\transparent[1]{%
    \errmessage{(Inkscape) Transparency is used (non-zero) for the text in Inkscape, but the package 'transparent.sty' is not loaded}%
    \renewcommand\transparent[1]{}%
  }%
  \providecommand\rotatebox[2]{#2}%
  \ifx\svgwidth\undefined%
    \setlength{\unitlength}{757.74481201bp}%
    \ifx\svgscale\undefined%
      \relax%
    \else%
      \setlength{\unitlength}{\unitlength * \real{\svgscale}}%
    \fi%
  \else%
    \setlength{\unitlength}{\svgwidth}%
  \fi%
  \global\let\svgwidth\undefined%
  \global\let\svgscale\undefined%
  \makeatother%
  \begin{picture}(1,0.35361832)%
    \put(0,0){\includegraphics[width=\unitlength,page=1]{robot2.pdf}}%
    \put(0.42714022,0.23339033){\color[rgb]{0,0,0}\makebox(0,0)[lb]{\smash{Instruments}}}%
    \put(0.42631374,0.28239671){\color[rgb]{0,0,0}\makebox(0,0)[lb]{\smash{Endoscope}}}%
    \put(0,0){\includegraphics[width=\unitlength,page=2]{robot2.pdf}}%
    \put(0.59412749,0.29777128){\color[rgb]{0,0,0}\makebox(0,0)[lb]{\smash{Origin (lens)}}}%
    \put(0,0){\includegraphics[width=\unitlength,page=3]{robot2.pdf}}%
    \put(0.94637007,0.09939316){\color[rgb]{0,0,0}\makebox(0,0)[lb]{\smash{Yaw}}}%
    \put(0,0){\includegraphics[width=\unitlength,page=4]{robot2.pdf}}%
    \put(0.80199429,0.09345448){\color[rgb]{0,0,0}\makebox(0,0)[lb]{\smash{Pitch}}}%
    \put(0,0){\includegraphics[width=\unitlength,page=5]{robot2.pdf}}%
    \put(0.87431415,0.0100491){\color[rgb]{0,0,0}\makebox(0,0)[lb]{\smash{Roll}}}%
    \put(0.641109,0.03417331){\color[rgb]{0,0,0}\makebox(0,0)[lb]{\smash{Roll=0, Pitch=0}}}%
    \put(0.97738315,0.06097653){\color[rgb]{0.35686275,0.60784314,0.83529412}\makebox(0,0)[lb]{\smash{Z}}}%
    \put(0.8853205,0.13626573){\color[rgb]{0.43921569,0.67843137,0.27843137}\makebox(0,0)[lb]{\smash{Y}}}%
    \put(0.90301775,0.0862225){\color[rgb]{1,0,0}\makebox(0,0)[lb]{\smash{X}}}%
    \put(0.76092506,0.33819386){\color[rgb]{0.43921569,0.67843137,0.27843137}\makebox(0,0)[lb]{\smash{Y}}}%
    \put(0.82379583,0.29862909){\color[rgb]{1,0,0}\makebox(0,0)[lb]{\smash{X}}}%
    \put(0.75864197,0.23690647){\color[rgb]{0.35686275,0.60784314,0.83529412}\makebox(0,0)[lb]{\smash{Z}}}%
    \put(0,0){\includegraphics[width=\unitlength,page=6]{robot2.pdf}}%
    \put(0.88460786,0.20535231){\color[rgb]{0,0,0}\makebox(0,0)[lb]{\smash{Tip points}}}%
  \end{picture}%
\endgroup%

%% file: figs/realistic_render2.pdf_tex
%% Creator: Inkscape inkscape 0.92.1, www.inkscape.org
%% Accompanies image file 'realistic_render2.pdf' (pdf, eps, ps)
\begingroup%
  \makeatletter%
  \providecommand\color[2][]{%
    \errmessage{(Inkscape) Color is used for the text in Inkscape, but the package 'color.sty' is not loaded}%
    \renewcommand\color[2][]{}%
  }%
  \providecommand\transparent[1]{%
    \errmessage{(Inkscape) Transparency is used (non-zero) for the text in Inkscape, but the package 'transparent.sty' is not loaded}%
    \renewcommand\transparent[1]{}%
  }%
  \providecommand\rotatebox[2]{#2}%
  \ifx\svgwidth\undefined%
    \setlength{\unitlength}{1092.75bp}%
    \ifx\svgscale\undefined%
      \relax%
    \else%
      \setlength{\unitlength}{\unitlength * \real{\svgscale}}%
    \fi%
  \else%
    \setlength{\unitlength}{\svgwidth}%
  \fi%
  \global\let\svgwidth\undefined%
  \global\let\svgscale\undefined%
  \makeatother%
  \begin{picture}(1,0.24776939)%
    \put(0,0){\includegraphics[width=\unitlength,page=1]{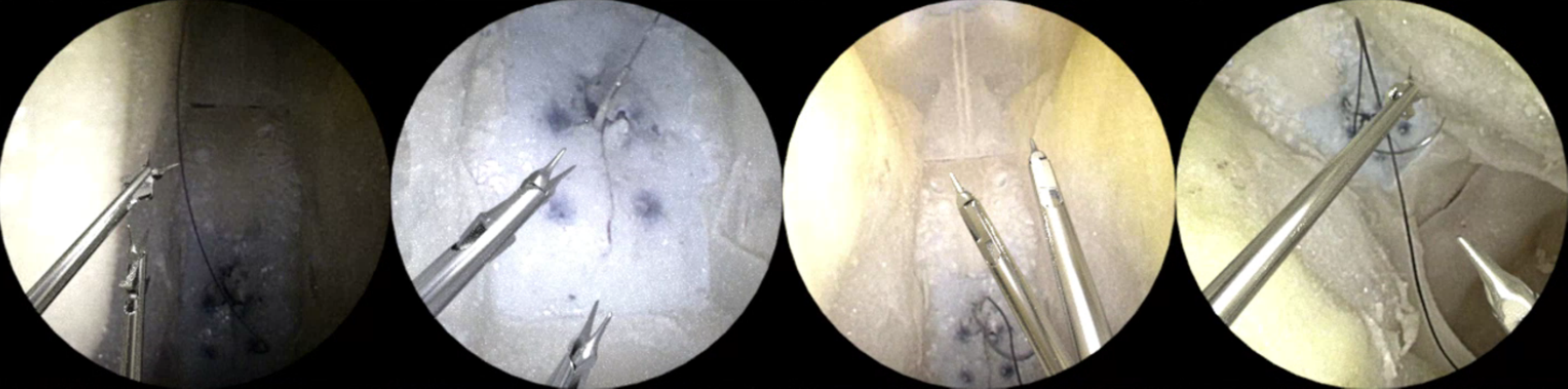}}%
  \end{picture}%
\endgroup%

%% file: figs/dr_components.pdf_tex
%% Creator: Inkscape 1.0.2 (e86c8708, 2021-01-15), www.inkscape.org
%% Accompanies image file 'dr_components.pdf' (pdf, eps, ps)
\begingroup%
  \makeatletter%
  \providecommand\color[2][]{%
    \errmessage{(Inkscape) Color is used for the text in Inkscape, but the package 'color.sty' is not loaded}%
    \renewcommand\color[2][]{}%
  }%
  \providecommand\transparent[1]{%
    \errmessage{(Inkscape) Transparency is used (non-zero) for the text in Inkscape, but the package 'transparent.sty' is not loaded}%
    \renewcommand\transparent[1]{}%
  }%
  \providecommand\rotatebox[2]{#2}%
  \newcommand*\fsize{\dimexpr\f@size pt\relax}%
  \newcommand*\lineheight[1]{\fontsize{\fsize}{#1\fsize}\selectfont}%
  \ifx\svgwidth\undefined%
    \setlength{\unitlength}{799.69454956bp}%
    \ifx\svgscale\undefined%
      \relax%
    \else%
      \setlength{\unitlength}{\unitlength * \real{\svgscale}}%
    \fi%
  \else%
    \setlength{\unitlength}{\svgwidth}%
  \fi%
  \global\let\svgwidth\undefined%
  \global\let\svgscale\undefined%
  \makeatother%
  \begin{picture}(1,0.42346167)%
    \lineheight{1}%
    \setlength\tabcolsep{0pt}%
    \put(0,0){\includegraphics[width=\unitlength,page=1]{dr_components.pdf}}%
    \put(0.33141974,0.32065993){\color[rgb]{0,0,0}\makebox(0,0)[lt]{\lineheight{1.25}\smash{\begin{tabular}[t]{l}cubic\end{tabular}}}}%
    \put(0.32691803,0.30561667){\color[rgb]{0,0,0}\makebox(0,0)[lt]{\lineheight{1.25}\smash{\begin{tabular}[t]{l}texture\end{tabular}}}}%
    \put(0.38911677,0.36598973){\color[rgb]{0,0,0}\makebox(0,0)[lt]{\lineheight{1.25}\smash{\begin{tabular}[t]{l}augmentation\end{tabular}}}}%
    \put(0.38441497,0.26591403){\color[rgb]{0,0,0}\makebox(0,0)[lt]{\lineheight{1.25}\smash{\begin{tabular}[t]{l}non-contextual\end{tabular}}}}%
    \put(0.4085742,0.2509083){\color[rgb]{0,0,0}\makebox(0,0)[lt]{\lineheight{1.25}\smash{\begin{tabular}[t]{l}images\end{tabular}}}}%
    \put(0.69292023,0.2658765){\color[rgb]{0,0,0}\makebox(0,0)[lt]{\lineheight{1.25}\smash{\begin{tabular}[t]{l}tool\end{tabular}}}}%
    \put(0.74135127,0.36975367){\color[rgb]{0,0,0}\makebox(0,0)[lt]{\lineheight{1.25}\smash{\begin{tabular}[t]{l}per part\end{tabular}}}}%
    \put(0.74165136,0.26591403){\color[rgb]{1,0,0}\makebox(0,0)[lt]{\lineheight{1.25}\smash{\begin{tabular}[t]{l}metallic\end{tabular}}}}%
    \put(0.74270176,0.24903258){\color[rgb]{1,0,0}\makebox(0,0)[lt]{\lineheight{1.25}\smash{\begin{tabular}[t]{l}scratch\end{tabular}}}}%
    \put(0.73542402,0.16387507){\color[rgb]{1,0,0}\makebox(0,0)[lt]{\lineheight{1.25}\smash{\begin{tabular}[t]{l}non-\end{tabular}}}}%
    \put(0.7343939,0.13592388){\color[rgb]{1,0,0}\makebox(0,0)[lt]{\lineheight{1.25}\smash{\begin{tabular}[t]{l}convex\end{tabular}}}}%
    \put(-0.00113723,0.21534472){\color[rgb]{0,0,0}\makebox(0,0)[lt]{\lineheight{1.25}\smash{\begin{tabular}[t]{l}light\end{tabular}}}}%
    \put(0.05226816,0.36704014){\color[rgb]{0,0,0}\makebox(0,0)[lt]{\lineheight{1.25}\smash{\begin{tabular}[t]{l}position\end{tabular}}}}%
    \put(0.05046748,0.25904891){\color[rgb]{0,0,0}\makebox(0,0)[lt]{\lineheight{1.25}\smash{\begin{tabular}[t]{l}intensity\end{tabular}}}}%
    \put(0.05331856,0.15477159){\color[rgb]{0,0,0}\makebox(0,0)[lt]{\lineheight{1.25}\smash{\begin{tabular}[t]{l}number\end{tabular}}}}%
    \put(0.06307229,0.04864358){\color[rgb]{0,0,0}\makebox(0,0)[lt]{\lineheight{1.25}\smash{\begin{tabular}[t]{l}kind\end{tabular}}}}%
    \put(0.33770964,0.10221403){\color[rgb]{0,0,0}\makebox(0,0)[lt]{\lineheight{1.25}\smash{\begin{tabular}[t]{l}tool\end{tabular}}}}%
    \put(0.42046624,0.15589702){\color[rgb]{0,0,0}\makebox(0,0)[lt]{\lineheight{1.25}\smash{\begin{tabular}[t]{l}color\end{tabular}}}}%
    \put(0.40486028,0.05226996){\color[rgb]{0,0,0}\makebox(0,0)[lt]{\lineheight{1.25}\smash{\begin{tabular}[t]{l}glossiness\end{tabular}}}}%
    \put(0.0507984,0.38279034){\color[rgb]{0,0,0}\makebox(0,0)[lt]{\lineheight{1.25}\smash{\begin{tabular}[t]{l}dr1:\end{tabular}}}}%
    \put(0.05026409,0.27740868){\color[rgb]{0,0,0}\makebox(0,0)[lt]{\lineheight{1.25}\smash{\begin{tabular}[t]{l}dr2:\end{tabular}}}}%
    \put(0.04948945,0.17330766){\color[rgb]{0,0,0}\makebox(0,0)[lt]{\lineheight{1.25}\smash{\begin{tabular}[t]{l}dr3:\end{tabular}}}}%
    \put(0.04948945,0.06826756){\color[rgb]{0,0,0}\makebox(0,0)[lt]{\lineheight{1.25}\smash{\begin{tabular}[t]{l}dr4:\end{tabular}}}}%
    \put(0.39102197,0.3829126){\color[rgb]{0,0,0}\makebox(0,0)[lt]{\lineheight{1.25}\smash{\begin{tabular}[t]{l}dr5:\end{tabular}}}}%
    \put(0.39102197,0.28349965){\color[rgb]{0,0,0}\makebox(0,0)[lt]{\lineheight{1.25}\smash{\begin{tabular}[t]{l}dr6:\end{tabular}}}}%
    \put(0.39289769,0.17095667){\color[rgb]{0,0,0}\makebox(0,0)[lt]{\lineheight{1.25}\smash{\begin{tabular}[t]{l}dr7:\end{tabular}}}}%
    \put(0.39289769,0.069668){\color[rgb]{0,0,0}\makebox(0,0)[lt]{\lineheight{1.25}\smash{\begin{tabular}[t]{l}dr8:\end{tabular}}}}%
    \put(0.73990517,0.38666403){\color[rgb]{0,0,0}\makebox(0,0)[lt]{\lineheight{1.25}\smash{\begin{tabular}[t]{l}dr9:\end{tabular}}}}%
    \put(0.74178089,0.28349965){\color[rgb]{0,0,0}\makebox(0,0)[lt]{\lineheight{1.25}\smash{\begin{tabular}[t]{l}dr10:\end{tabular}}}}%
    \put(0.73990517,0.18408669){\color[rgb]{0,0,0}\makebox(0,0)[lt]{\lineheight{1.25}\smash{\begin{tabular}[t]{l}dr11:\end{tabular}}}}%
    \put(0.73439627,0.15008236){\color[rgb]{1,0,0}\makebox(0,0)[lt]{\lineheight{1.25}\smash{\begin{tabular}[t]{l}contextual\end{tabular}}}}%
  \end{picture}%
\endgroup%

%% file: figs/model.pdf_tex
%% Creator: Inkscape inkscape 0.92.1, www.inkscape.org
%% Accompanies image file 'model.pdf' (pdf, eps, ps)
\begingroup%
  \makeatletter%
  \providecommand\color[2][]{%
    \errmessage{(Inkscape) Color is used for the text in Inkscape, but the package 'color.sty' is not loaded}%
    \renewcommand\color[2][]{}%
  }%
  \providecommand\transparent[1]{%
    \errmessage{(Inkscape) Transparency is used (non-zero) for the text in Inkscape, but the package 'transparent.sty' is not loaded}%
    \renewcommand\transparent[1]{}%
  }%
  \providecommand\rotatebox[2]{#2}%
  \ifx\svgwidth\undefined%
    \setlength{\unitlength}{939.81344604bp}%
    \ifx\svgscale\undefined%
      \relax%
    \else%
      \setlength{\unitlength}{\unitlength * \real{\svgscale}}%
    \fi%
  \else%
    \setlength{\unitlength}{\svgwidth}%
  \fi%
  \global\let\svgwidth\undefined%
  \global\let\svgscale\undefined%
  \makeatother%
  \begin{picture}(1,0.45519806)%
    \put(0,0){\includegraphics[width=\unitlength,page=1]{model.pdf}}%
    \put(0.80316779,0.09228291){\color[rgb]{0,0,0}\rotatebox{90.00000252}{\makebox(0,0)[lb]{\smash{Box Aware Head }}}}%
    \put(0.04829247,0.14525087){\color[rgb]{0,0,0}\makebox(0,0)[lb]{\smash{ResNet34}}}%
    \put(0,0){\includegraphics[width=\unitlength,page=2]{model.pdf}}%
    \put(0.84280332,0.24124864){\color[rgb]{0,0,0}\makebox(0,0)[lb]{\smash{Bounding Box}}}%
    \put(0.84918756,0.18858925){\color[rgb]{0,0,0}\makebox(0,0)[lb]{\smash{Pose}}}%
    \put(0.84982599,0.13545105){\color[rgb]{0,0,0}\makebox(0,0)[lb]{\smash{Class}}}%
    \put(0,0){\includegraphics[width=\unitlength,page=3]{model.pdf}}%
    \put(0.74080435,0.04264753){\color[rgb]{0,0,0}\makebox(0,0)[lb]{\smash{= }}}%
    \put(0.76206389,0.04264753){\color[rgb]{0,0,0}\makebox(0,0)[lb]{\smash{(Conv+Bn+Relu)$\times$2}}}%
    \put(0.74164495,0.0081726){\color[rgb]{0,0,0}\makebox(0,0)[lb]{\smash{= }}}%
    \put(0.7581163,0.0081726){\color[rgb]{0,0,0}\makebox(0,0)[lb]{\smash{Conv+Sigmoid}}}%
    \put(0.0697563,0.37194479){\color[rgb]{0,0,0}\makebox(0,0)[lb]{\smash{Input}}}%
    \put(0.29180036,0.41375096){\color[rgb]{0,0,0}\makebox(0,0)[lb]{\smash{Segmentation}}}%
    \put(0.48911611,0.09142316){\color[rgb]{0,0,0}\makebox(0,0)[lb]{\smash{Extended ResNet18}}}%
    \put(0.22174391,0.02045589){\color[rgb]{0.92941176,0.49019608,0.19215686}\makebox(0,0)[lb]{\smash{Proposed Double Encoder}}}%
    \put(0.4779756,0.41239962){\color[rgb]{0.92941176,0.49019608,0.19215686}\makebox(0,0)[lb]{\smash{Proposed Mask Guidance}}}%
  \end{picture}%
\endgroup%

%% file: figs/polaris_setup.pdf_tex
%% Creator: Inkscape inkscape 0.92.1, www.inkscape.org
%% Accompanies image file 'polaris_setup.pdf' (pdf, eps, ps)
\begingroup%
  \makeatletter%
  \providecommand\color[2][]{%
    \errmessage{(Inkscape) Color is used for the text in Inkscape, but the package 'color.sty' is not loaded}%
    \renewcommand\color[2][]{}%
  }%
  \providecommand\transparent[1]{%
    \errmessage{(Inkscape) Transparency is used (non-zero) for the text in Inkscape, but the package 'transparent.sty' is not loaded}%
    \renewcommand\transparent[1]{}%
  }%
  \providecommand\rotatebox[2]{#2}%
  \ifx\svgwidth\undefined%
    \setlength{\unitlength}{670.03518677bp}%
    \ifx\svgscale\undefined%
      \relax%
    \else%
      \setlength{\unitlength}{\unitlength * \real{\svgscale}}%
    \fi%
  \else%
    \setlength{\unitlength}{\svgwidth}%
  \fi%
  \global\let\svgwidth\undefined%
  \global\let\svgscale\undefined%
  \makeatother%
  \begin{picture}(1,0.53209144)%
    \put(0,0){\includegraphics[width=\unitlength,page=1]{polaris_setup.pdf}}%
    \put(0.01333545,0.42144054){\color[rgb]{0,0,0}\makebox(0,0)[lb]{\smash{Optical marker}}}%
    \put(0,0){\includegraphics[width=\unitlength,page=2]{polaris_setup.pdf}}%
    \put(0.01333545,0.29732766){\color[rgb]{0,0,0}\makebox(0,0)[lb]{\smash{Instrument}}}%
    \put(0,0){\includegraphics[width=\unitlength,page=3]{polaris_setup.pdf}}%
    \put(0.01333545,0.21942131){\color[rgb]{0,0,0}\makebox(0,0)[lb]{\smash{Optical marker }}}%
    \put(0.01333545,0.18718419){\color[rgb]{0,0,0}\makebox(0,0)[lb]{\smash{for an endoscope}}}%
  \end{picture}%
\endgroup%

%% file: figs/hand_eye_proposed.pdf_tex
%% Creator: Inkscape inkscape 0.92.1, www.inkscape.org
%% Accompanies image file 'hand_eye_proposed.pdf' (pdf, eps, ps)
\begingroup%
  \makeatletter%
  \providecommand\color[2][]{%
    \errmessage{(Inkscape) Color is used for the text in Inkscape, but the package 'color.sty' is not loaded}%
    \renewcommand\color[2][]{}%
  }%
  \providecommand\transparent[1]{%
    \errmessage{(Inkscape) Transparency is used (non-zero) for the text in Inkscape, but the package 'transparent.sty' is not loaded}%
    \renewcommand\transparent[1]{}%
  }%
  \providecommand\rotatebox[2]{#2}%
  \ifx\svgwidth\undefined%
    \setlength{\unitlength}{928.26544189bp}%
    \ifx\svgscale\undefined%
      \relax%
    \else%
      \setlength{\unitlength}{\unitlength * \real{\svgscale}}%
    \fi%
  \else%
    \setlength{\unitlength}{\svgwidth}%
  \fi%
  \global\let\svgwidth\undefined%
  \global\let\svgscale\undefined%
  \makeatother%
  \begin{picture}(1,0.52510791)%
    \put(0,0){\includegraphics[width=\unitlength,page=1]{hand_eye_proposed.pdf}}%
    \put(0.2242725,0.45751943){\color[rgb]{0,0,0}\makebox(0,0)[lb]{\smash{annotation}}}%
    \put(0.07916315,0.19013902){\color[rgb]{0,0,0}\makebox(0,0)[lb]{\smash{Project the CAD model }}}%
    \put(0.00967871,0.16683749){\color[rgb]{0,0,0}\makebox(0,0)[lb]{\smash{corresponding to $\Delta$X and $\Delta$Z}}}%
    \put(0.47074294,0.00729046){\color[rgb]{0,0,0}\makebox(0,0)[lb]{\smash{Update $\Delta$X and $\Delta$Z}}}%
    \put(0.8622581,0.28212779){\color[rgb]{0,0,0}\makebox(0,0)[lb]{\smash{IOU}}}%
    \put(0.85403847,0.05928205){\color[rgb]{0,0,0}\makebox(0,0)[lb]{\smash{Edge}}}%
    \put(0.63155901,0.30872578){\color[rgb]{0,0,0}\makebox(0,0)[lb]{\smash{Compute }}}%
    \put(0.60143831,0.28545658){\color[rgb]{0,0,0}\makebox(0,0)[lb]{\smash{correspondence}}}%
  \end{picture}%
\endgroup%

%% file: figs/dr_eval.pdf_tex
%% Creator: Inkscape inkscape 0.92.1, www.inkscape.org
%% Accompanies image file 'dr_eval.pdf' (pdf, eps, ps)
\begingroup%
  \makeatletter%
  \providecommand\color[2][]{%
    \errmessage{(Inkscape) Color is used for the text in Inkscape, but the package 'color.sty' is not loaded}%
    \renewcommand\color[2][]{}%
  }%
  \providecommand\transparent[1]{%
    \errmessage{(Inkscape) Transparency is used (non-zero) for the text in Inkscape, but the package 'transparent.sty' is not loaded}%
    \renewcommand\transparent[1]{}%
  }%
  \providecommand\rotatebox[2]{#2}%
  \ifx\svgwidth\undefined%
    \setlength{\unitlength}{732.09640503bp}%
    \ifx\svgscale\undefined%
      \relax%
    \else%
      \setlength{\unitlength}{\unitlength * \real{\svgscale}}%
    \fi%
  \else%
    \setlength{\unitlength}{\svgwidth}%
  \fi%
  \global\let\svgwidth\undefined%
  \global\let\svgscale\undefined%
  \makeatother%
  \begin{picture}(1,0.47802911)%
    \put(0,0){\includegraphics[width=\unitlength,page=1]{dr_eval.pdf}}%
    \put(0.01257353,0.08556572){\color[rgb]{0.34901961,0.34901961,0.34901961}\makebox(0,0)[lb]{\smash{-30}}}%
    \put(0.01257353,0.16371116){\color[rgb]{0.34901961,0.34901961,0.34901961}\makebox(0,0)[lb]{\smash{-20}}}%
    \put(0.01257353,0.24187027){\color[rgb]{0.34901961,0.34901961,0.34901961}\makebox(0,0)[lb]{\smash{-10}}}%
    \put(0.03696922,0.32002937){\color[rgb]{0.34901961,0.34901961,0.34901961}\makebox(0,0)[lb]{\smash{0}}}%
    \put(0.02329616,0.39817481){\color[rgb]{0.34901961,0.34901961,0.34901961}\makebox(0,0)[lb]{\smash{10}}}%
    \put(0.08808271,0.05969481){\color[rgb]{0.34901961,0.34901961,0.34901961}\makebox(0,0)[lb]{\smash{\textbf{+}\textbf{d}\textbf{r}\textbf{1}}}}%
    \put(0.16868685,0.05969481){\color[rgb]{0.34901961,0.34901961,0.34901961}\makebox(0,0)[lb]{\smash{\textbf{+}\textbf{ }\textbf{d}\textbf{r}\textbf{2}}}}%
    \put(0.25635289,0.05969481){\color[rgb]{0.34901961,0.34901961,0.34901961}\makebox(0,0)[lb]{\smash{\textbf{+}\textbf{d}\textbf{r}\textbf{3}}}}%
    \put(0.34050848,0.05969481){\color[rgb]{0.34901961,0.34901961,0.34901961}\makebox(0,0)[lb]{\smash{\textbf{+}\textbf{d}\textbf{r}\textbf{4}}}}%
    \put(0.4246504,0.05969481){\color[rgb]{0.34901961,0.34901961,0.34901961}\makebox(0,0)[lb]{\smash{\textbf{+}\textbf{d}\textbf{r}\textbf{5}}}}%
    \put(0.50880598,0.05969481){\color[rgb]{0.34901961,0.34901961,0.34901961}\makebox(0,0)[lb]{\smash{\textbf{+}\textbf{d}\textbf{r}\textbf{6}}}}%
    \put(0.59296156,0.05969481){\color[rgb]{0.34901961,0.34901961,0.34901961}\makebox(0,0)[lb]{\smash{\textbf{+}\textbf{d}\textbf{r}\textbf{7}}}}%
    \put(0.67711714,0.05969481){\color[rgb]{0.34901961,0.34901961,0.34901961}\makebox(0,0)[lb]{\smash{\textbf{+}\textbf{d}\textbf{r}\textbf{8}}}}%
    \put(0.76127272,0.05969481){\color[rgb]{0.34901961,0.34901961,0.34901961}\makebox(0,0)[lb]{\smash{\textbf{+}\textbf{d}\textbf{r}\textbf{9}}}}%
    \put(0.83833907,0.05969481){\color[rgb]{0.34901961,0.34901961,0.34901961}\makebox(0,0)[lb]{\smash{\textbf{+}\textbf{d}\textbf{r}\textbf{1}\textbf{0}}}}%
    \put(0.92249465,0.05969481){\color[rgb]{0.34901961,0.34901961,0.34901961}\makebox(0,0)[lb]{\smash{\textbf{+}\textbf{d}\textbf{r}\textbf{1}\textbf{1}}}}%
    \put(0.00379054,0.45076351){\color[rgb]{0,0,0}\makebox(0,0)[lb]{\smash{Performance gain [\%]}}}%
    \put(0.1386225,0.00970139){\color[rgb]{0.34901961,0.34901961,0.34901961}\makebox(0,0)[lb]{\smash{Translation}}}%
    \put(0.31863977,0.00970139){\color[rgb]{0.34901961,0.34901961,0.34901961}\makebox(0,0)[lb]{\smash{Orientation}}}%
    \put(0.49733208,0.00970139){\color[rgb]{0.34901961,0.34901961,0.34901961}\makebox(0,0)[lb]{\smash{mAP}}}%
    \put(0.6024685,0.00970139){\color[rgb]{0.34901961,0.34901961,0.34901961}\makebox(0,0)[lb]{\smash{Detected rate}}}%
    \put(0.81000947,0.00970139){\color[rgb]{0.34901961,0.34901961,0.34901961}\makebox(0,0)[lb]{\smash{Average}}}%
  \end{picture}%
\endgroup%

%% file: figs/result_visualization4.pdf_tex
%% Creator: Inkscape inkscape 0.92.1, www.inkscape.org
%% Accompanies image file 'result_visualization4.pdf' (pdf, eps, ps)
\begingroup%
  \makeatletter%
  \providecommand\color[2][]{%
    \errmessage{(Inkscape) Color is used for the text in Inkscape, but the package 'color.sty' is not loaded}%
    \renewcommand\color[2][]{}%
  }%
  \providecommand\transparent[1]{%
    \errmessage{(Inkscape) Transparency is used (non-zero) for the text in Inkscape, but the package 'transparent.sty' is not loaded}%
    \renewcommand\transparent[1]{}%
  }%
  \providecommand\rotatebox[2]{#2}%
  \ifx\svgwidth\undefined%
    \setlength{\unitlength}{639.24000549bp}%
    \ifx\svgscale\undefined%
      \relax%
    \else%
      \setlength{\unitlength}{\unitlength * \real{\svgscale}}%
    \fi%
  \else%
    \setlength{\unitlength}{\svgwidth}%
  \fi%
  \global\let\svgwidth\undefined%
  \global\let\svgscale\undefined%
  \makeatother%
  \begin{picture}(1,0.77407547)%
    \put(0,0){\includegraphics[width=\unitlength,page=1]{result_visualization4.pdf}}%
  \end{picture}%
\endgroup%